\documentclass[conference]{IEEEtran}

\usepackage{amsmath,amssymb,latexsym,amsfonts,mathrsfs}
\DeclareMathOperator*{\argmax}{arg\,max}

\ifCLASSINFOpdf

\else

\fi

\usepackage{graphicx}
\begin{document}

\title{Improving Image Clustering using Sparse Text and the Wisdom of the Crowds}

\author{
\IEEEauthorblockN{Anna Ma}
\IEEEauthorblockA{Institute of\\Mathematical Sciences\\
Claremont Graduate University\\
Claremont, CA 91711\\
anna.ma@cgu.edu}
\and
\IEEEauthorblockN{Arjuna Flenner}
\IEEEauthorblockA{Physics and\\Computational Sciences\\
Naval Air Warfare Center\\
China Lake, CA 93555\\
arjuna.flenner@navy.mil}
\and
\IEEEauthorblockN{Deanna Needell}
\IEEEauthorblockA{Department of\\Mathematical Sciences\\
Claremont McKenna College\\ 
Claremont, CA 91711 \\
dneedell@cmc.edu}
\and
\IEEEauthorblockN{Allon G.\ Percus}
\IEEEauthorblockA{Institute of\\Mathematical Sciences\\
Claremont Graduate University\\
Claremont, CA 91711\\
allon.percus@cgu.edu}
}

\maketitle

\begin{abstract} We propose a method to improve image clustering using sparse text and the wisdom of the crowds. In particular, we present a method to fuse two different kinds of document features, image and text features, and use a common dictionary or ``wisdom of the crowds'' as the connection between the two different kinds of documents. With the proposed fusion matrix, we use topic modeling via non-negative matrix factorization to cluster documents.

\end{abstract}

\IEEEpeerreviewmaketitle

\section{Introduction} There has been substantial research in organizing large image databases. Often, these images have corresponding text, such as captions in textbooks and metatags. We investigate strategies to use this text information to improve clustering image documents into groups of similar images. Image Clustering is used for image database management, content based image searches, and image classification. In this paper, we present a method for improving image clusters using sparse text and freely obtainable information form the internet.  The motivation behind our method stems from the idea that we can fuse image and text documents and use the ``wisdom of the crowds'' (WOC), the freely obtainable information, to connect the sparse text documents where WOC documents act as a representative of a single class.

In Section 2, we breifly touch upon related material. In Section 3, we introduce our method of fusing text and image documents using the term frequency-inverse document frequency weighting scheme. We then describe how non-negative matrix factorization is used for the purpose of topic modeling in section 4. In Section 5, we present results from an application of our method. 

\section{Related Works}
There have been many studies on text document clustering and image clustering. A general joint image and text clustering strategy proceeds in two steps, first two different types of documents must be combined into a single document feature matrix. Then, a clustering technique is implemented.

The term frequency-inverse document frequency (TF-IDF) is a technique to create a feature matrix from a collection, or corpus, of documents.  TF-IDF is a weighting scheme that weighs features in documents based on how often the words occurs in an individual document compared with how often it occurs in other documents \cite{tf-idf_original}.  TF-IDF has been used for text mining, near duplicate detection, and information retrieval. When dealing with text documents, the natural features to use are words (i.e. delimiting strings by white space to obtain features). We can represent each word by a unique integer.  

In order to use text processing techniques for image databases, we generate a collection of image words using two steps.  First, we obtain a collection of image features, and then define a mapping from the image features to the integers.  To obtain image features, we use the scale invariant feature transform (SIFT) \cite{lowe2004distinctive}.   We then use k-means to cluster the image features into $K$ different clusters.  The mapping from the image feature to the cluster is used to identify image words, and results in the image Bag-Of-Words model \cite{fei2005bayesian}. 

Topic modeling is used to uncover a hidden topical structure of a collection of documents. There have been studies on using large scale data collections to improve classification of sparse, short segments of text, which usually cluster inaccurately due to spareness of text \cite{phan2008learning}. Latent Dirichlet Allocation (LDA), singular value decomposition (SVD), and non-negative matrix factorization (NNMF) are just some of the models that have been used in topic modeling \cite{arora2012learning}.

In our method, we integrate these techniques to combine and cluster different types of documents. We use SIFT to obtain image features and term frequency-inverse document frequency to generate a feature matrix in the fused collection of documents. Then, we use the non-negative matrix factorization to learn a representation of the corpus which is used to cluster the documents.

\section{Fusing Image and Text Documents} We denote a collection of image documents $D = \{d_1, ... d_n\}$ and a collection of sparse text documents $S = \{s_1,...s_n\}$ and text document $s_i$ describes image document $d_i$ for i=1,...m.   Some of the text documents may be empty, indicating the absence of any labeled text.  

\textbf{Image Documents} Using the scale invariant feature transform (SIFT) and k-means, we obtain $A \in \mathbb{R}^{n \times p}$ where $p$ is the number of image features and n is the number of image documents and element $A_{i,j}$ represents the number of times the image document $d_i$ contains the $j^{th}$ feature.
  
\textbf{Wisdom of the Crowds} Due to the sparse nature of the text documents we are considering, the WOC is needed to link features that represent a single class. For example, if one wishes to obtain a class of documents and images about cats, text and images from a wikipedia page on cats can be used as the wisdom of the crowds. Using Wikipedia, we collect WOC documents $W = \{w_1, ... w_k\}$ where $k$ is the number of clusters we wish to cluster the images into. Each $w_i$ is a text document that contains features that collectively describe a single class. To create text features, we parse text documents by white space (i.e. break up text by words) and obtain a corpus $f = (f_1, ... f_q)$ of $q$ unique features. Let $C \in \mathbb{R}^{k \times q}$ Each $C_{i,j}$ is the number of times the feature $f_j$ appears in $w_i$.

\textbf{Text Documents} In the same manner as with WOC documents, we parse text documents into features to obtain a corpus. In most cases, the features in this corpus have already appeared somewhere in the WOC documents so we use the same $f = (f_1,...f_q)$ from the previous step. If it is not the case, ``missing" features can simply be appended to the list of features and the $C$ matrix extended to reflect the absence of the missing features. We calculate $B \in \mathbb{R}^{m \times q}$ where $m$ is the number of text documents, $q$ is the number of features in corpus $f$, and element $B_{i,j}$ is the number of times text document $s_i$ contains the feature $f_j$. We then extend $B \in \mathbb{R}^{m \times q}$ to $B \in \mathbb{R}^{n \times q}$ such that $B_{i,j}= 0$ for $i = m+1,...n, j=1,...q$. Intuitively, this means that none of the text features knowingly describes the $m+1, ... n$ image documents.

We combine the image feature matrix $A$, the text feature matrix $B$, and the WOC matrix $C$ to initialize matrix $M$:
 $$M = \begin{bmatrix}
	A & B\\
	0 & C\\
\end{bmatrix},$$
where $M \in \mathbb{R}^{(m+k) \times (p+q)}$ and $0 \in \{0\}^{k \times p}$. We call $M$ our mixed document feature matrix. Each row represents a document and each column represents a feature (either an image feature or a text feature). 

Without the reweighing using IDF, it is difficult to use sparse text to aid in image classification. This is because the frequency of the image features outweigh any sort of effect the sparse text has in the classification of image documents. The inverse document frequency matrix $\text{IDF} \in \mathbb{R}^{p+q \times p+q}$ is defined as the diagonal matrix with nonzero elements:
$$ \text{IDF}_{j,j} = \log\dfrac{m+k}{|\{i : M_{i,j}>0\}|}, $$
where ${|\{i : M_{i,j}>0\}|}$ is the number of documents containing the $j^{th}$ feature. We then re-evaluate M to be $M = M \times IDF$. 

\begin{figure}[htp]
\centering
\includegraphics[scale=.450]{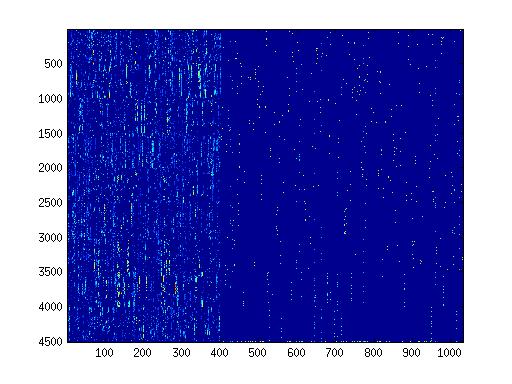}
\caption{Example of an M matrix with 4500 image documents, 9 WOC documents, and 450 text documents.}
\label{fig:mixmat}
\end{figure}

\section{Topic Modeling using Non-negative Matrix Factorization} We use non-negative matrix factorization (NNMF) on the document feature matrix to cluster documents into topics. We consider the document feature matrix as a set of (m+k) points in a (p+q) dimensional space. Each document is a point and each feature is a dimension. We want to reduce the dimensionality of this space into $k^* << \min(m+k, p+q)$ dimensions \cite{Lsas}.
NMF is a method that takes a non-negative matrix $M_+ \in \mathbb{R}^{(m+k) \times (p+q)}$ and factors it into two non-negative matrices $U_+ \in \mathbb{R}^{(m+k) \times k^*}$ and $V_+ \in \mathbb{R}^{k^* \times (p+q)}$ where $k^*$ is the rank of the desired lower dimensional approximation to $X$ \cite{LeeSeung}. We take the $(p+q)$-dimensional feature space and project it onto a $k^*$-dimensional topic space where $k^*$ is the number of desired classes.
Denoting the Frobenius norm of $M$ as $||M||_F^2 = \sum_i\sum_j M_{i,j}^2$, we wish to obtain $U$ and $V$ by minimizing the following cost function:
\begin{equation}
||M - UV||^2_F.
\label{eq:nmf}
\end{equation}
Intuitavely, $U_{i,j}$ tells us how well document $d_i$ fits into topic $j$ and $V_{i,j}$ tells us how well the $j^{th}$ feature describes the $i^{th}$ topic. In most applications of topic modeling using NNMF, a document $d_i$ belongs to topic $j$ if 
 $$j = \argmax_z U_{i,z}.$$
Because of the geometric nature of the NNMF topic modeling method, we also investigate the clusters that result from a k-means clustering on the rows of $U$, or the location of documents in the reduced-dimension topical space.

\section{Results}
\subsection{Evaluation Metrics} Purity and z-Rand scores are metrics used to evaluate cluster quality \cite{traud2008comparing}, \cite{amigo2009comparison}. 
 
\textbf{Purity} Purity is a well known clustering measure that depends on some ground truth. This metric compares a cluster to the ground truth by comparing the intersection of the ground truth clustering with the new clustering. Purity can be computed by as follows:
\begin{equation*}
Purity(G,C) = \frac{1}{m} \sum_i \max_j |g_j \cap c_i|.
\end{equation*}
Here, $m$ is the number of documents, $G = \{g_1, ..., g_k\}$ is the ground truth or class assignment where each $g_j$ is a set of indices belonging to the $j^{th}$ class, and $C = \{c_1, ... c_t\}$ is the clustering from some method where each $c_i$ is the set of indices belonging to the $i^{th}$ cluster. It is important to note that purity is sensitive to the number of clusters. If every document had its own cluster, then the purity for this set of clusters is 1. To address this sensitivity, we also look at the $z$-rand metric.

\textbf{Z-rand} To define the $z$-rand score we first define $p$ to be the number of pairs of documents that are in the same cluster as determined by our method and in the ground truth (i.e. the number of document pairs that are correctly clustered together). The $z$-rand score, $z_R$ is defined as:
$$z_R = \dfrac{(p-\mu_p)}{\sigma_p},$$ 
where $\mu_p$ and $\sigma_p$ are the expected value the standard deviation of $p$ under a hypergeometric distribution with the same size of clusters. Intuitively, we are comparing the number of correctly identified pairings to the number of correctly identified pairings if the pairings were randomly selected. The higher the z-rand score, the better clusters as the clusters created are very different from randomly picked clusters.\\

We apply our method to the Electro-Optical (EO) dataset provided by China Lake. This dataset consists of 9 classes of images where each class contains 500 images of a single vehicle from different angles. Because this dataset does not contain text data, we use wikipedia articles to create sparse text captions for a varying number of documents by randomly selecting 5 words from each wikipedia article to be an image caption. Using only the image documents and NNMF, the clusters produced score a mean purity of 0.6397 and mean z-rand of 1460.7.

\begin{table}[h!]
\begin{center}
\begin{tabular}{|c|c|c|} \hline
	Matrix & Purity & Zrand\\ \hline
	A & 0.6397 $\pm$ 0.012 & 1460.7 $\pm$ 52.18\\ \hline
	$[A : B]$ & 0.6597 $\pm$ 0.01& 1538.6 $\pm$ 45.71\\ \hline
	M & 0.769 $\pm$ 0.0012 & 1909.5 $\pm$ 136.55
	\\ \hline
\end{tabular}
\end{center} 
\caption{Results from the EO data set. A is only using image features, $[A:B]$ is image features with sparse text, and M is image and text features with additional dicionary.}
\label{tab:mainresults} 
 \end{table}

For our first experiment, we investigate the usefulness of fusing image and text documents together and using the appropriate reweighting. In Table \ref{tab:mainresults}, we are comparing using only image features, using image and sparse text features, and using image features, sparse text features, and a dictionary. As one can see, using only image features, does the worst while using sparse text features helps only slightly. We attribute this slight improve to the fact that the text documents are sparse. When we use the WOC, we get a significant increase in purity and zrand.

We also investigated the effect of varying the percentage of documents with both image and text features and found that in general, regardless of the number of image documents that contained sparse text, the purity stayed from 0.76-0.78, while the z-rand ranged from 1877.0-1938.2. To improve results, one may also remove stop words from the text features. Stop words are commonly used words such as `the', `a', and `is'. When we did this, we obtained a mean purity of 0.778.

We found that each class can be broken down into three subclasses: front of vehicle, back of vehicle, and sides. So, using $k=27$, we greatly improve our results as shown in Table \ref{tab:27topics}. 
\begin{table}[h!]
\begin{center}
\begin{tabular}{|c|c|c|}
\hline
\% of documents with labels & purity & z-Rand \\\hline
0.2&0.88126$\pm$0.0035533&1579.9675$\pm$15.6499\\\hline
0.4&0.87793$\pm$0.0046333&1571.9551$\pm$14.8006\\\hline
0.6&0.88403$\pm$0.0043387&1566.322$\pm$13.8568\\\hline
0.8&0.88341$\pm$0.0042595&1580.0351$\pm$13.7404\\\hline
1&0.88071$\pm$0.0038168&1576.8284$\pm$16.8794\\\hline
\end{tabular}
\end{center}
\caption{Purity and z-Rand over different percentages of images documents with text documents where the number of text documents $m = \lfloor{np}\rfloor$ for EO data set using $k^* = 27$.}
\label{tab:27topics}
\end{table}

\section{Conclusion}
Fusing text documents and image documents makes it possible to improve image clusters. The results from the EO data set show that are method does make an improvement on the image clusters when comparing to using NNMF on only the image document feature matrix $A$.  
\section*{Acknowledgements}
This work was supported in part by AFOSR MURI grant FA9550-10-1-0569. 
Arjuna Flenner was supported by ONR grants number N0001414WX20237 and N0001414WX20170. Deanna Needell was partially supported by Simons Foundation Collaboration grant $\#274305$.

\bibliography{asilomar_imgtxt.bib}{}

\begin{thebibliography}{1}

\bibitem{amigo2009comparison}
Enrique Amig{\'o}, Julio Gonzalo, Javier Artiles, and Felisa Verdejo.
\newblock A comparison of extrinsic clustering evaluation metrics based on
  formal constraints.
\newblock {\em Information retrieval}, 12(4):461--486, 2009.

\bibitem{arora2012learning}
Sanjeev Arora, Rong Ge, and Ankur Moitra.
\newblock Learning topic models--going beyond svd.
\newblock In {\em Foundations of Computer Science (FOCS), 2012 IEEE 53rd Annual
  Symposium on}, pages 1--10. IEEE, 2012.

\bibitem{fei2005bayesian}
Li~Fei-Fei and Pietro Perona.
\newblock A bayesian hierarchical model for learning natural scene categories.
\newblock In {\em Computer Vision and Pattern Recognition, 2005. CVPR 2005.
  IEEE Computer Society Conference on}, volume~2, pages 524--531. IEEE, 2005.

\bibitem{Lsas}
Hyunsoo Kim and Haesun Park.
\newblock Nonnegative matrix factorization based on alternating nonnegativity
  constrained least squares and active set method.
\newblock {\em SIAM Journal on Matrix Analysis and Applications},
  30(2):713--730, 2008.

\bibitem{LeeSeung}
Daniel~D. Lee and H.~Sebastian Seung.
\newblock {\em Algorithms for Non-negative Matrix Factorization }.

\bibitem{lowe2004distinctive}
David~G Lowe.
\newblock Distinctive image features from scale-invariant keypoints.
\newblock {\em International journal of computer vision}, 60(2):91--110, 2004.

\bibitem{phan2008learning}
Xuan-Hieu Phan, Le-Minh Nguyen, and Susumu Horiguchi.
\newblock Learning to classify short and sparse text \& web with hidden topics
  from large-scale data collections.
\newblock In {\em Proceedings of the 17th international conference on World
  Wide Web}, pages 91--100. ACM, 2008.

\bibitem{tf-idf_original}
Gerard Salton and Chris Buckley.
\newblock Term weighting approaches in automatic text retrieval.
\newblock Technical report, Ithaca, NY, USA, 1987.

\bibitem{traud2008comparing}
Amanda~L Traud, Eric~D Kelsic, Peter~J Mucha, and Mason~A Porter.
\newblock Comparing community structure to characteristics in online collegiate
  social networks.
\newblock {\em arXiv preprint arXiv:0809.0690}, 2008.

\end{thebibliography}
\bibliographystyle{plain}
\end{document}